\documentclass[letterpaper, 10 pt, conference]{ieeeconf}  

\IEEEoverridecommandlockouts                              

\usepackage{graphicx} 
\usepackage{epsfig} 
\usepackage{mathptmx} 
\usepackage{times} 
\usepackage{amsmath} 
\usepackage{amssymb}  
\usepackage[font=footnotesize]{caption}
\usepackage{float}
\usepackage{hyperref}
\usepackage{pifont}

\usepackage[ruled,vlined]{algorithm2e}

\usepackage[capitalize]{cleveref}

\let\oldtwocolumn\twocolumn
\renewcommand\twocolumn[1][]{%
    \oldtwocolumn[{#1}{
    \begin{center}
           \includegraphics[width=\textwidth]{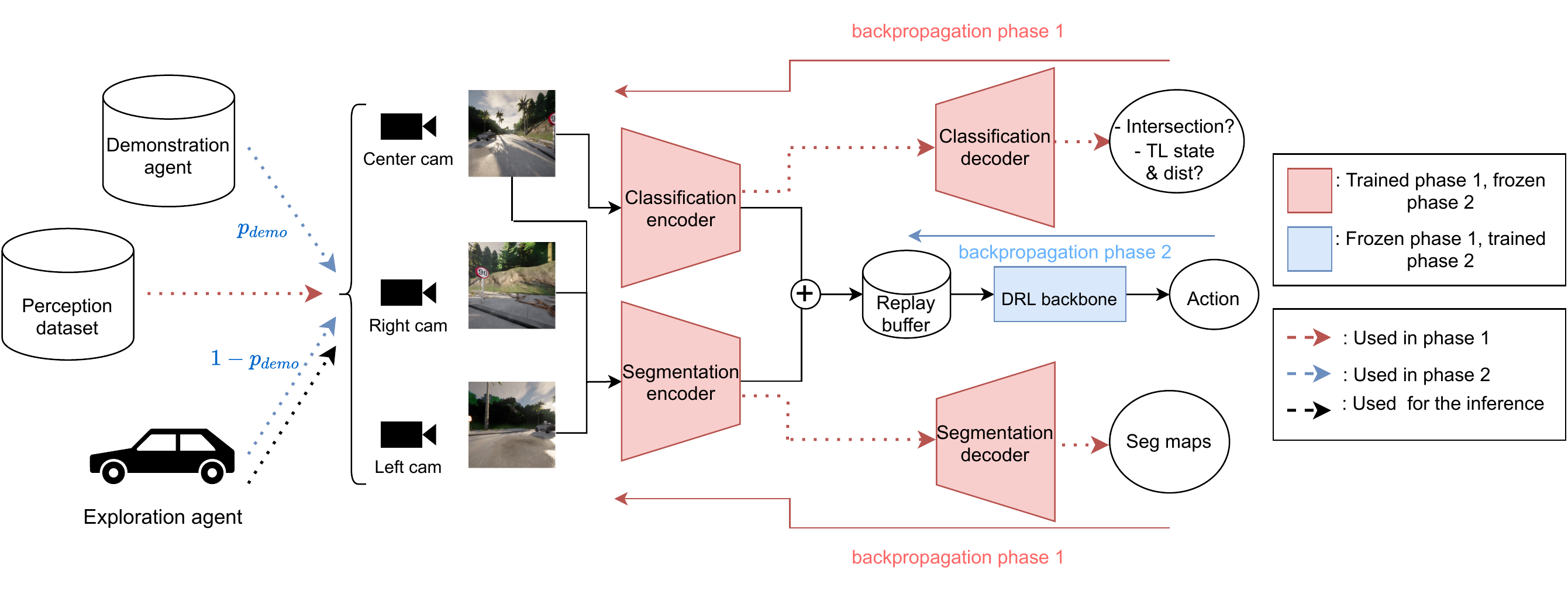}
           \captionof{figure}{General Reinforced Imitation (GRI) is applied to visual-based autonomous driving in an end-to-end pipeline composed of a perception module encoding RGB images from the three cameras on the driving agent and a decision-making module inferring an action from the encoded features. This pipeline is trained in two phases: (1) Visual encoders are pretrained on a perception dataset on several auxiliary tasks, which are semantic segmentation, road type classification, relevant traffic light presence, and if there is such a traffic light, its state and the distance to it. (2) Visual encoders are frozen and a GRI-based DRL network is trained with both pre-generated expert data with an offline demonstration agent and an online exploration agent gathering data from a simulator. At any given training step, the next episode to add to the replay buffer comes from the demonstration agent with a probability of $p_{demo}$, else from the exploration agent. Actions correspond to a pair (steering, throttle) to apply to the car.}
           \label{fig:full_griad}
        \end{center}
    }]
}

\begin{document}

\title{\LARGE \bf GRI: General Reinforced Imitation 
\\and its Application to Camera-Based Autonomous Driving}

\author{Raphaël Chekroun$^{1,2}$, Marin Toromanoff$^2$, Sascha Hornauer$^1$, Fabien Moutarde$^1$\\
$^{1}$Mines Paris - PSL University, $^{2}$Valeo DAR\\
\texttt{firstname.lastname@\{minesparis.psl.eu, valeo.com\}} 
}

\maketitle
\thispagestyle{empty}
\pagestyle{empty}


\begin{abstract}
    Deep reinforcement learning (DRL) has been demonstrated to be effective for several complex decision-making applications such as autonomous driving and robotics. However, DRL is notoriously limited by its high sample complexity and its lack of stability. Prior knowledge, e.g. as expert demonstrations, is often available but challenging to leverage to mitigate these issues. In this paper, we propose General Reinforced Imitation (GRI), a novel method which combines benefits from exploration and expert data and is straightforward to implement over any off-policy RL algorithm. We make one simplifying hypothesis: expert demonstrations can be seen as perfect data whose underlying policy gets a constant high reward. Based on this assumption, GRI introduces the notion of offline demonstration agents. This agent sends expert data which are processed both concurrently and indistinguishably with the experiences coming from the online RL exploration agent. We show that our approach enables major improvements on camera-based autonomous driving in urban environments. We further validate the GRI method on Mujoco continuous control tasks with different off-policy RL algorithms. Our method ranked first on the CARLA Leaderboard and outperforms World on Rails, the previous state-of-the-art, by 17\%.
\end{abstract}
\section{INTRODUCTION}
\label{sec:intro}

Autonomous driving (AD) in urban areas is a convoluted task. Agents have to efficiently analyse a highly complex environment and make online decisions to follow driving rules whilst simultaneously interacting with other dynamic agents, such as drivers or pedestrians. That is why, literature in autonomous driving focuses on different learning methods rather than the design of general hand-crafted rules.

Imitation learning (IL) \cite{bojarski, il-bullshit, transfuser, marin-fisheye}, especially behavior cloning, aims at mimicking expert behavior for a given task. It requires a significant amount of annotated data, often recorded by human drivers. Even though this kind of data can be recorded easily on a large scale, practical safety concerns in real traffic lead to heavily biased observations showing predominantly safe driving examples, and under-represents rare dangerous situations. Hence, IL agents suffer from distribution mismatch and will struggle to recover from its own mistakes.

Deep reinforcement learning (DRL) \cite{dqn, ppo, td3, a3c} offers an alternative, more robust to distribution mismatch than IL, by letting the agent learn from its own mistakes through trial-and-error. In the RL framework, the agent explores its environment by itself and gathers rewards, a numerical value assessing how much a given action in a given state is good. The goal of the agent is to maximize its cumulative rewards. To do so, the agent needs to optimize sequences of actions rather than instantaneous ones.
Nonetheless, DRL needs an order of magnitude more data than IL to converge due to this extensive, and often time-consuming, exploration of the environment during the training. 

To overcome IL distribution mismatch and RL data inefficiency, we propose General Reinforced Imitation (GRI), a novel method which combines both exploration and prior knowledge from demonstrations. GRI is based on the simplifying hypothesis that expert data presents a perfect behavior and therefore, an expert's action should receive a constant, high reward. Straightforward to implement over any off-policy algorithm, GRI introduces the notion of an offline \textit{demonstration agent}. This offline agent sends expert data associated with a constant demonstration reward to the replay buffer of an RL online exploration agent. We note that those expert data are processed by the DRL algorithm concurrently and indistinguishably from the exploration data


The GRI method is applied to visual-based autonomous driving in an end-to-end pipeline on the CARLA simulator \cite{carla}, an open-source simulator for research in autonomous driving. We call this algorithm GRI for Autonomous Driving (GRIAD). The whole pipeline is represented in \cref{fig:full_griad}. On the CARLA Leaderboard, an online benchmark ranking agents according to the quality of their driving, we achieved 17\% better results than World on Rails \cite{wor}, the prior top ranking entry. In addition, our method used only three cameras and no LiDAR, which is fewer sensors than the other top entries \cite{wor, transfuser}. At the time of writing (February 2022), GRIAD is the best camera-based agent on the CARLA Leaderboard according to the main metric, the driving score. We also conducted ablation studies to highlight the impact of GRIAD compared to standard RL training on the CARLA NoCrash Benchmark \cite{nocrash}.

Finally, we conducted experiments on the Mujoco \cite{mujoco} benchmark to investigate our method adaptability and generalizability. Tests were conducted on four different Mujoco environments, with two different DRL algorithm as backbones. Our experiments demonstrate that using the GRI framework systematically leads to better results, even when the expert data are noisy or not significantly better than the trained vanilla RL algorithm.

We summarize our main contributions below:
\begin{itemize}
\itemsep0em 

    \item Definition of the novel GRI method to combine offline demonstrations and online exploration.
    \item Presentation and ablation study of GRI for visual-based Autonomous Driving (GRIAD) algorithm.
    \item Further analysis of GRI-based algorithms on the Mujoco benchmark. 
\end{itemize}

\section{RELATED WORK}
The GRI method aims at leveraging both offline expert demonstrations and online simulator exploration. Our main application is end-to-end camera-based autonomous driving on the CARLA simulator \cite{carla}. Therefore, this section focuses both on end-to-end autonomous driving methods that achieved milestones on CARLA, and existing decision-making methods learning from demonstration and exploration.
\label{sec:bib}
\subsection{End-to-end Autonomous Driving on CARLA}
End-to-end autonomous driving, i.e. directly mapping sensor signals to control is a highly complex task on which training an agent with DRL is tedious. 
IL methods were the first to lead the CARLA Leaderboard. In particular, Learning by Cheating (LBC) \cite{lbc} presents an efficient method to train a behavior cloning agent in two steps: (i) train a privileged  behavior cloning agent which has access to all the ground truth data, and (ii) train a behavior cloning agent to mimic the privileged one. An evaluation of several methods on the NoCrash benchmark, presented in Chen et al. \cite{wor}, shows that LBC presents great results on the training conditions but generalizes poorly on unknown environments.

DRL can also be used for end-to-end autonomous driving. However, camera-based DRL comes with some drawbacks. Indeed, image inputs are often of high dimensions thus requiring larger DRL networks which are usually difficult to train to convergence. Therefore, for camera-based DRL, one can encode the sensors' signal in a more compact and semantically rich representation to train the DRL network on this predefined latent space as in D. Gordon et al. \cite{splitnet}. This latent space can be obtained by pretraining a visual encoder on some visual tasks, such as segmentation or classification.


Based on this principle, Toromanoff et al. \cite{architecture-marin} introduced the \textit{Implicit Affordances} (IAs) method. They design and train an efficient DRL agent on CARLA, winning the CARLA challenge two years in a row. To do so, they propose an end-to-end pipeline composed of two subsystems trained successively. First, a visual encoder is trained on some auxiliary tasks. Those tasks are semantic segmentation, classification of the type of road, detection of traffic lights, and if there is a relevant traffic light, the state of, and distance to, the light. Then, the visual encoder is frozen and the DRL-based decision-making subsystem is trained on the encoder latent space.

Another top ranked camera-based agent on the CARLA Leaderboard is World on Rails \cite{wor} which assumes the world to be on rails, meaning that the agent's actions affect only its own state and do not influence its environment. Based on that hypothesis, they transpose the driving problem into a simple, yet powerful, tabular RL setup.

Finally, Transfuser \cite{transfuser} and Learning from All Vehicles, more recent top ranked agents on the CARLA Leaderboard, mainly focuses on LiDAR and camera fusion.

Other concurrent work combines an RL driving coach and an IL learner, mediated with a learned bird's map  \cite{roach} but are not currently in the CARLA Leaderboard.

\subsection{Learning from demonstration and exploration}


The aforementioned IL and RL strengths and weaknesses are complementary. Indeed, IL suffers from distribution mismatch contrarily to online RL. Alternatively, as RL learns from scratch it is less data efficient than IL which incorporates prior demonstration knowledge during training.

To take the best of both worlds, some algorithms combine IL and RL to maximize efficacy by leveraging both expert data and exploration \cite{dqfd, sqil, dapg, sacr2}. 
In particular, demonstrations can be used to initialize policies by pretraining the network \cite{nac, dapg, dqfd} or leveraged with a specifically designed reward \cite{dqfd, sqil}.





Soft-Q Imitation Learning (SQIL) \cite{sqil} and Deep Q-learning from Demonstrations (DQfD) \cite{dqfd} are the two closest approaches to ours as both take advantage of demonstrations in a different way and can be applied to any off-policy RL algorithms.

SQIL \cite{sqil} does IL using an RL agent. To do so, the replay buffer is initially filled with demonstrations, associated with a constant reward $r_{demo}=1$. A RL agent collects data from exploration into the replay buffer, associated with a constant reward $r_{explo} = 0$. Thus, SQIL designed an RL agent that learns to imitate expert behavior, and has been mathematically demonstrated to be equivalent to regularized behavior cloning. 
However, SQIL does not efficiently leverage exploration as environment rewards are never used. Our method combines both the IL part from SQIL and the classical, rich RL online exploration.

DQfD \cite{dqfd} is based on DQN \cite{dqn}, an off-policy RL algorithm with a replay buffer. DQfD first pretrains the agent on expert data with both IL and RL losses using the real reward given by the environment. After some steps of pretraining, the agent starts gathering data from the environment in the memory buffer. The network is then trained on batches composed of exploration data with a RL loss and expert data with both IL and RL losses. 
Nonetheless, DQfD uses simultaneously reinforcement and imitation which can have divergent losses and are difficult to jointly optimize \cite{rl-id}. Our method leverages demonstrations and exploration exclusively with an RL loss, and thus cannot suffer from divergent losses issue. Moreover, DQfD, contrarily to GRI, relies on the true environment reward for the expert data, which cannot always be obtained.

\section{GENERAL REINFORCED IMITATION (GRI)}
Our pipeline for autonomous driving is an end-to-end system. Its decision-making subsystem uses the GRIAD algorithm, an adaptation of the GRI method to visual-based autonomous driving (AD) on CARLA. This section presents the GRI method and details the whole pipeline.
\subsection{Method}
\label{sec:gri}
GRI is a method which is straightforward to implement over any off-policy RL algorithm using a replay buffer, such as SAC \cite{sac}, DDPG \cite{ddpg}, DQN \cite{dqn} and its successive improvements \cite{rainbow, iqn}. GRI is built upon the hypothesis that expert demonstrations can be seen as perfect data whose underlying policy gets a constant high reward. We denote this as demonstration reward, $r_{demo}$. In our experiments we chose $r_{demo}$ to be the maximum of the reward.

\begin{algorithm}
\SetAlgoLined
Input: $r_{demo}$ demonstration reward value, $p_{demo}$ probability to use demonstration agent\;
 Initialize empty buffer $\mathcal{B}$\;

 \While{not converged}{
   \If{len($\mathcal{B}$) $\geq$ min\_buffer}{
    do a DRL network update\;
    }
 
    \eIf{random.random() $\geq$ $p_{demo}$}{
        collect episode $(s^{\mbox{\textit{online}}}_t, a_t, r_t, s^{\mbox{\textit{online}}}_{t+1})_t$ in buffer $\mathcal{B}$ with exploration agent\
    }
    {
        add episode $(s^{\mbox{\textit{offline}}}_t, a_t, r_{demo}, s^{\mbox{\textit{offline}}}_{t+1})_t$ in buffer $\mathcal{B}$ with demonstration agent\;
    }
  

 }
 \caption{GRI: General Reinforced Imitation}
 \label{algo:gri}
\end{algorithm}

The idea of GRI is to distill expert knowledge from demonstrations into an RL agent during the training phase. To do so, we defined two types of agents: (i) the online \textit{exploration agent}, which is the regular RL agent exploring its environment to gather experiences $(s^{\mbox{\textit{online}}}_t, a_t, r_t, s^{\mbox{\textit{online}}}_{t+1})$ into the memory buffer, and (ii) the offline \textit{demonstration agent} which sends expert data associated with a constant demonstration reward $(s^{\mbox{\textit{offline}}}_t, a_t, r_{demo}, s^{\mbox{\textit{offline}}}_{t+1})$ to the memory buffer. $s_t$ is the state, $a_t$ the chosen action and $r_t$ the reward at time $t$. At any given training step, the next episode to add to the replay buffer comes from the demonstration agent with a probability of $p_{demo}$, else from the exploration agent. GRI is summarized in \cref{algo:gri}. 

\subsection{GRI for Autonomous Driving}
We applied GRI on a pipeline inspired by Toromanoff et al. Implicit Affordances method \cite{architecture-marin}. 
As this method is trained in two phases, hence making it modular, we optimized both the visual and the decision-making subsystems independently. 
\newline

\textbf{Design of the vision subsystem} We first train two visual encoders on segmentation and classifications tasks with different camera perspectives to extract compact semantic features.

We found that single camera setup is leading to more collisions on intersections, as sensors are not able to see close obstacles while turning. Thus, we mounted three RGB cameras on the hood of our agent vehicle, at the coordinates $x=2.5m, z=1.2m$ and $y\in \{-0.8, 0, 0.8\}m$ relatively to the center of the car. The side cameras are angled at 70°. All three cameras have a 100° field of view. 



\begin{figure}[h!]
    \centering
    \includegraphics[width=1.0\linewidth]{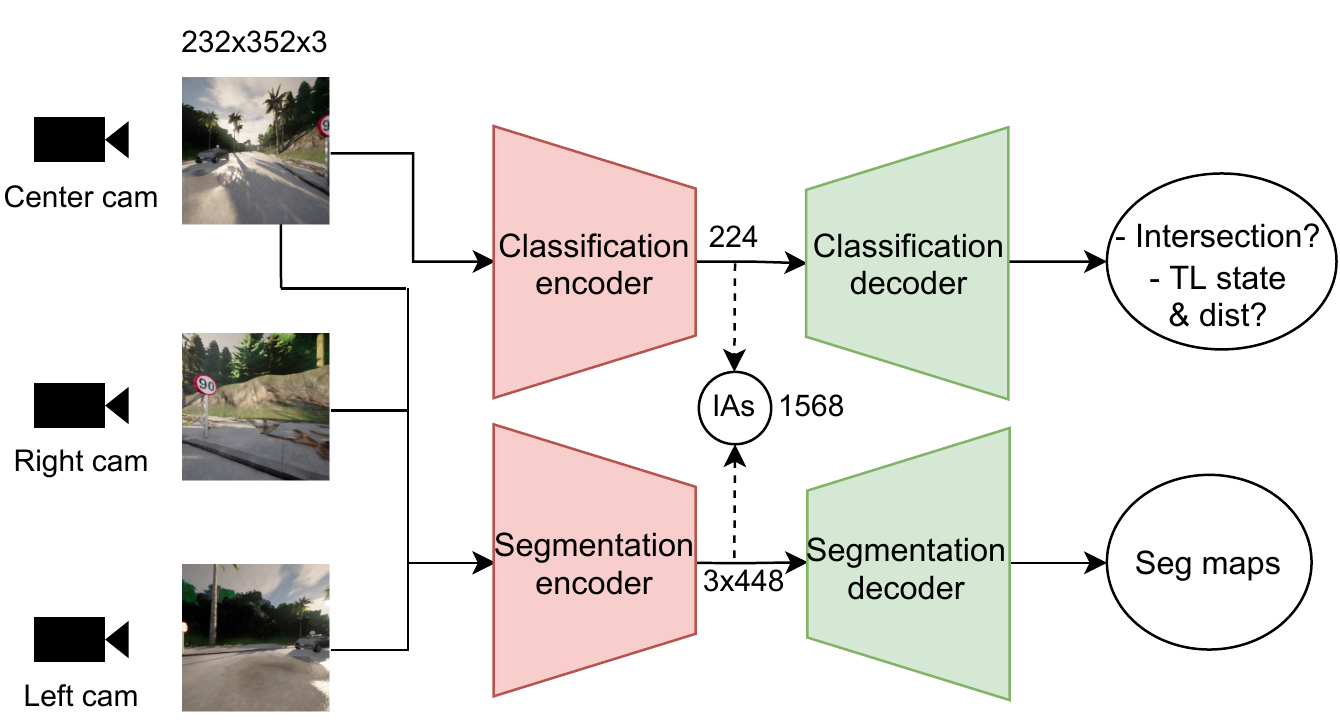}
    \caption{Feature extraction from RGB camera images for the visual subsystem. Two encoder-decoder networks are pretrained on segmentation, classifications and regression tasks. Classifications and regression are only performed on the center image while all three images are segmented. After training, the visual encoders serve as fixed feature extractors with frozen weights. For the DRL backbone training, both encoder outputs are concatenated and sent to the memory buffer as input to DRL. Both encoders are Efficientnet-b1. The segmentation decoder is fully convolutional, and the classification decoder is an MLP with several outputs.}
    \label{fig:encoder}
\end{figure}

Our visual subsystem is composed of two highly specialized Efficientnet-b1 \cite{efficientnet} models, one for the segmentation and one for the classification and regression tasks, as shown in \cref{fig:encoder}. We concatenate the four outputs (three segmentations, one for each camera, and one classification from the front camera only) of both Efficientnet-b1 and use it similarly as Implicit Affordances (IAs) for the DRL training. 

This architecture allows to keep the same accuracy on classification and segmentation metrics as if we were using a single bigger encoder for all the auxiliary tasks like in Toromanoff et al. \cite{architecture-marin}, while reducing the encoder latent space dimension by a factor of $\sim 5$. 


The visual part for the CARLA Leaderboard has been trained on a dataset of 400,000 samples which corresponds to 44 hours of driving. This dataset has been generated with the CARLA autopilot on every town with random trajectories. Each sample of the dataset is composed of three images from the three cameras and the corresponding ground truth information, which are segmentation maps from CARLA, booleans indicating the presence of an intersection and the presence of a traffic light in front of the car. And if there is a traffic light, a class corresponding to its color, and the distance to it in meters. Trajectories have been augmented with random cameras translations and rotations.
\newline

\textbf{Design of the Decision Subsystem} \label{sec:griad} The decision subsystem takes as input four consecutive encodings of the three camera images and outputs an action. Therefore, a state contains visual features from the last 300 milliseconds, as the simulator runs at 10 FPS. An action is defined by the combination of the desired steering of the wheel, and the throttle or brake to apply.

Generating data on the CARLA simulator is very computationally expensive. We used a Rainbow-IQN Ape-X \cite{drl-marin}, which is a distributed DRL backbone, to mitigate this issue.  

Due to Rainbow-IQN Ape-X being based on DQN \cite{dqn} the action state has to be discrete. Therefore, we discretized the action state in 27 steering values, and 4 throttle or brake values. The discretized action space contains $27\times 4 = 108$ actions.

\begin{figure}[h!]
    \centering
    \includegraphics[width=1.0\linewidth]{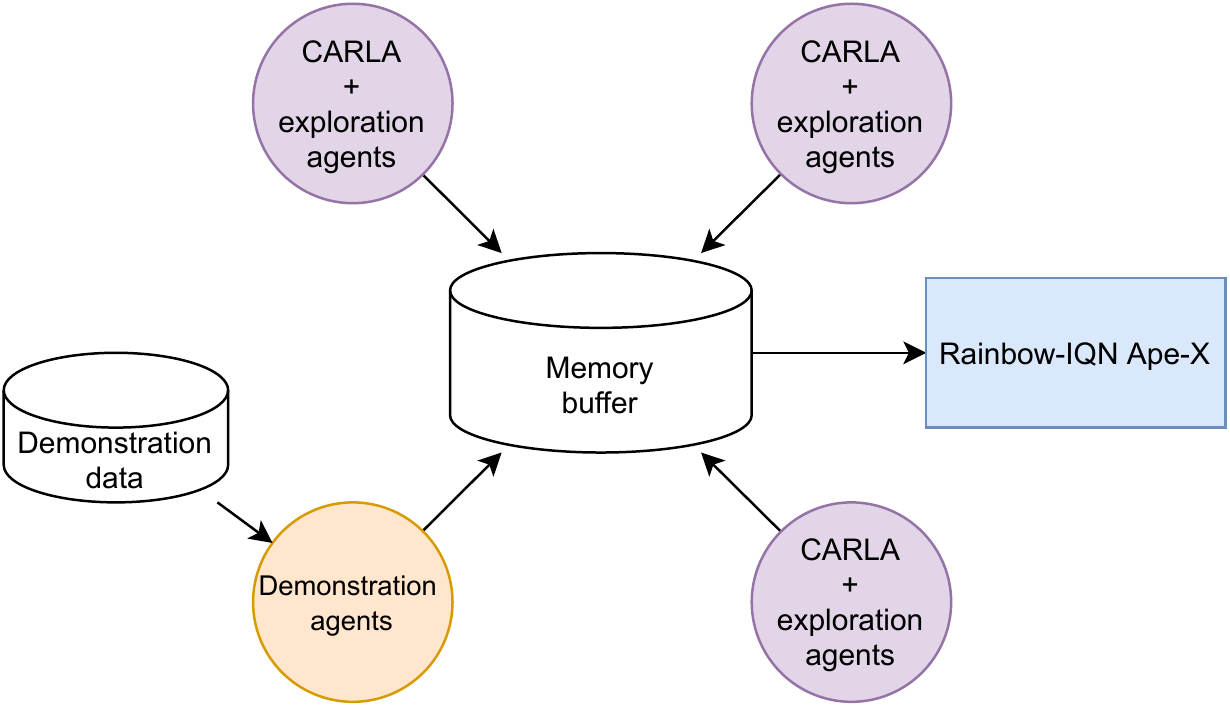}
    \caption{Simplified representation of the distributed GRIAD setup with a Rainbow-IQN Ape-X backbone. A central computer receives data in a shared replay buffer from both exploration and demonstration agents running on other computers. Data are sampled from this replay buffer to make the backpropagation and update the weights of all the agents. Images from the agents are encoded using the network presented in \cref{fig:encoder} before being stored in the memory buffer.}
    \label{fig:griad-setup}
\end{figure}

We called this setup GRI for Autonomous Driving (GRIAD). We diagram it in \cref{fig:griad-setup}.

The demonstration dataset contains 200,000 samples, which correspond to 22 hours of driving, generated using the autopilot from CARLA on predefined tracks published by the CARLA team\footnote{\url{https://github.com/carla-simulator/scenario_runner}}. Each sample from the demonstration dataset consists of three images from the three cameras and a discrete action obtained by mapping continuous actions of the expert to our discrete set of RL actions. We did not use any data augmentation. We note that the autopilot makes driving errors such as collisions, red light infractions, or the car getting stuck for hundreds of frames. As a result $\sim10\%$ of our demonstrations correspond to poor action choices. However, we decided to use this demonstration dataset as it is in order to assess the robustness of our method to noisy demonstrations.  

In our experiments on CARLA, GRIAD had a total of 12 agents, including 3 demonstration agents, running in a distributed setup and sending data to the memory buffer. As demonstration agents have been constrained to send data at the same frequency as exploration agents, this is equivalent to having $p_{demo}=25\%$.

The reward function used for the exploration agents is the same as in Toromanoff et al. \cite{architecture-marin}. Since this reward has a range between 0 and 1, we set the demonstration reward to $r_{demo} = 1$.

\section{EXPERIMENTAL RESULTS}
The GRI method was assessed on its primary application of visual-based autonomous driving on the CARLA Leaderboard and with an ablation study comparing it to vanilla RL. Further studies of the method have also been conducted on the Mujoco benchmark to analyze its behavior depending on the proportion of demonstration agents and highlight its generalizability to other DRL backbones. 

\subsection{GRIAD on CARLA}

\label{sec:exp}
\textbf{On the CARLA leaderboard.} We trained GRIAD for 60M steps ($\sim$45M exploration steps + 200,000 expert data sampled $\sim$15M times). Both visual and decision-making parts were trained on all available maps with all available weather. We compare to LBC \cite{lbc}, IAs \cite{architecture-marin}, Transfuser+ \cite{transfuser} and World on Rails \cite{wor}. 
Our method outperforms World on Rails, the previous leading method on the CARLA leaderboard, by $\sim 17\%$ on the main metric, the driving score, while using fewer sensors. However, more recent LiDAR-based methods give significantly better results, but cannot be compared directly as inputs are different. CARLA Leaderboard results are presented in \cref{Tab:leaderboard}.

\begin{table}[h!]
\centering
\begin{tabular}{|l|c|c|ccc|}
\hline

 Method  & \small{Cam.} & \small{LiDAR} & DS & RC & IS \\ \hline

\small{\textbf{GRIAD} (ours)}         & 3 & \ding{55} & \textbf{36.79}         & \textbf{61.85}            & \textbf{0.60}  \\
\small{Rails} \cite{wor}         & 4   &  \ding{55} & 31.37  & 57.65            & 0.56   \\
\small{IAs} \cite{architecture-marin}           & 1 &  \ding{55} & 24.98         & 46.97            & 0.52 \\     
\small{LBC} \cite{lbc}           & 3 & \ding{55} & 10.9           & 21.3          &  0.55         \\

\hline
\small{\textbf{LAV}}    & 4 &  \ding{51} & \textbf{61.8}         & \textbf{94.5}            & \textbf{0.64}   \\
\small{Transfuser+} \cite{transfuser}     & 4 &  \ding{51} & 50.5         & 73.8            & 0.68   \\

\hline
\end{tabular}
\caption{Comparison of camera-based and LiDAR-based agents on CARLA Leaderboard's driving metrics: driving score (DS, main metric), route completion (RC), and infraction score (IS). Results from the public CARLA Leaderboard website on February 2022. Higher is better for all metrics. Our method improves the driving score by 17\% relative to the prior camera-based state-of-the-art \cite{wor} while using fewer sensors than the two other best.}
\label{Tab:leaderboard}
\end{table}

\textbf{Ablation study on the NoCrash benchmark.} We compared GRIAD to regular RL by using the same architecture with and without demonstration agents on the NoCrash benchmark \cite{nocrash}. To do so, agents are trained on a single environment (Town01) under a specific set of training weather. Then, agents are evaluated on several scenarios with different traffic density on the training (Town01) and test (Town02) town with training and test sets of weather.
\begin{table}[h!]
\begin{tabular}{l|l|lll}
 \text{\footnotesize{Task}} & \text{\footnotesize{Town, Weather}} &  \text{\small{RL 12M}} & \text{\footnotesize{RL 16M}}& \text{\footnotesize{GRIAD}} \\

\hline \text{\footnotesize{Empty}} & & 96.3 $\pm$ 1.5 &  \textbf{98.0 $\pm$ 1.0} & \textbf{98.0 $\pm$ 1.7} \\
\text{\footnotesize{Regular}} & \text { train, train }& 95.0 $\pm$ 2.4 & \textbf{98.6 $\pm$ 1.2}  & \textbf{98.3 $\pm$ 1.7} \\
\text{\footnotesize{Dense}} & & 91.7 $\pm$ 2.0 & \textbf{95.0 $\pm$ 1.6} & 93.7 $\pm$ 1.7 \\

\hline \text{\footnotesize{Empty}} & & 83.3 $\pm$ 3.7 & \textbf{96.3 $\pm$ 1.7} & 94.0 $\pm$ 1.6\\
\text{\footnotesize{Regular}} & \text { test, train } & 82.6 $\pm$ 3.7 & \textbf{96.3 $\pm$ 2.5} & 93.0 $\pm$ 0.8 \\
\text{\footnotesize{Dense}} & & 61.6 $\pm$ 2.0 & \textbf{78.0 $\pm$ 2.8} & \textbf{77.7 $\pm$ 4.5}\\

\hline \text{\footnotesize{Empty}} & & 67.3 $\pm$ 1.9 & 73.3 $\pm$ 2.5 & \textbf{83.3 $\pm$ 2.5}\\
\text{\footnotesize{Regular}} & \text { train, test } & 76.7 $\pm$ 2.5 &  81.3 $\pm$ 2.5 & \textbf{86.7 $\pm$ 2.5} \\
\text{\footnotesize{Dense}} & & 67.3 $\pm$ 2.5 & 80.0 $\pm$ 1.6 & \textbf{82.6 $\pm$ 0.9}\\

\hline \text{\footnotesize{Empty}} & & 60.6 $\pm$ 2.5 & 62.0 $\pm$ 1.6 & \textbf{68.7 $\pm$ 0.9} \\
\text{\footnotesize{Regular}} & \text { test, test } & 59.3 $\pm$ 2.5 & 56.7 $\pm$ 3.4 & \textbf{63.3 $\pm$ 2.5}\\
\text{\footnotesize{Dense}} & & 40.0 $\pm$ 1.6 &  46.0 $\pm$ 3.3 & \textbf{52.0 $\pm$ 4.3}\\
\end{tabular}
\caption{Ablation study of GRIAD using the NoCrash benchmark. Mean and standard deviation over 3 evaluation seeds. Score is the percentage of road completed without any crash. GRIAD experimentally shows to generalize more on test weather than RL with 12M and 16M steps and globally gives the best agent. We note that, for computational reason, neither the RL nor GRIAD was trained until convergence. Hence, comparison to the state-of-the-art should rather be done on the CARLA Leaderboard with the fully trained GRIAD model, cf \cref{Tab:leaderboard}.}
\label{Tab:nocrash}
\end{table}

For these experiments, GRIAD was trained on 16M samples corresponding to 12M exploration steps + 25,000 expert data which have been sampled 4M times in total. We present an ablation study to show how GRIAD compares to RL without GRI i.e. without demonstration agents, using two vanilla RL models: one trained on 12M exploration steps and the other on 16M exploration steps. Each agent was trained using the exact same visual encoder trained on another demonstration dataset of 100,000 samples coming exclusively from Town01 under training weather. Results are presented in \cref{Tab:nocrash}.

We first observe that GRIAD systematically gives better results than RL with 12M steps, while taking approximately the same time to train (+$\sim 4\%$). Indeed, as demonstration agents do not require any interaction with the simulator, we can add them at a negligible cost and still improve results.

We also observe that while RL with 16M steps does better than GRIAD on train weather, GRIAD gives better results on the test weather while being $\sim 25\%$ faster to train. We believe this is because RL tends to overfit on a given environment if it explores it too much. Hence, replacing 4M exploration data with 25,000 demonstration data sampled $\sim 160$ times each appears to reduce the overfitting and allows a better generalization.

We also trained the same pipeline using the SQIL method during 20M steps, but the evaluation reward stayed particularly low during training. First test showed SQIL to be inefficient for autonomous driving on CARLA as it did not learn to drive at all, staying static or drifting off the road most of the time. It reached the score of 0 on every evaluated tasks. We believe that the reward signal as defined by SQIL is not rich enough to allow the network to converge on such a highly complex task.

\subsection{GRI on the Mujoco benchmark}
\label{sec:gri-mujoco}

\begin{figure}[!h]
    \centering
    \includegraphics[width=0.24\linewidth, height=0.2\linewidth]{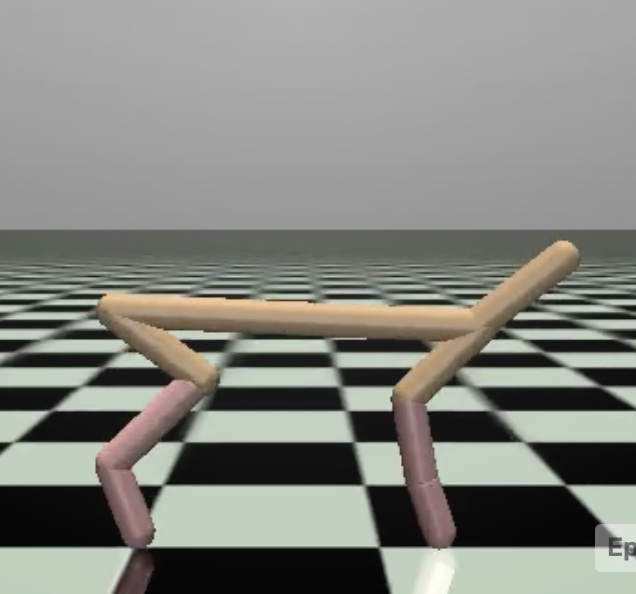}  
    \includegraphics[width=0.24\linewidth, height=0.2\linewidth]{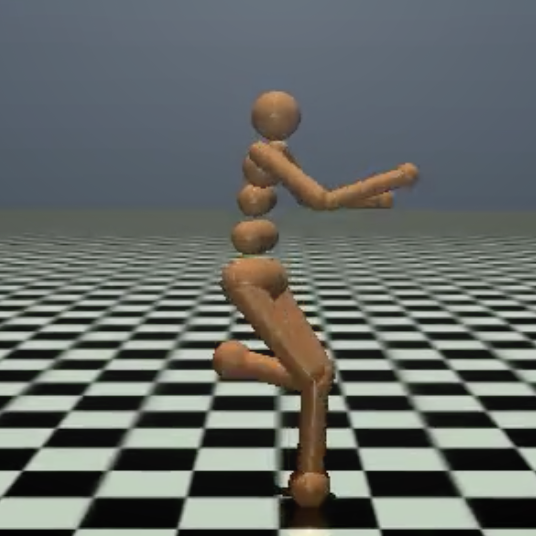}      \includegraphics[width=0.24\linewidth, height=0.2\linewidth]{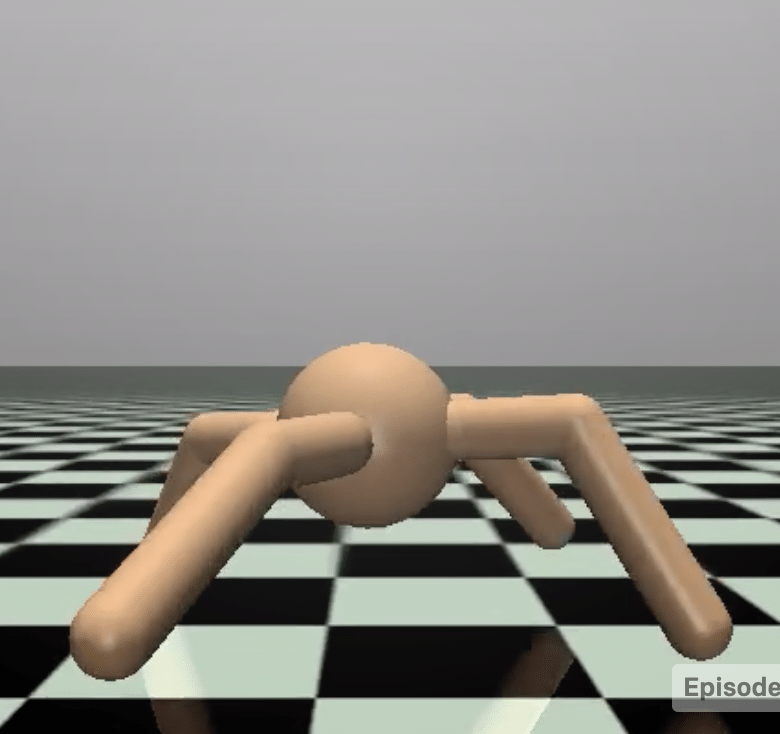}
    \includegraphics[width=0.24\linewidth, height=0.2\linewidth]{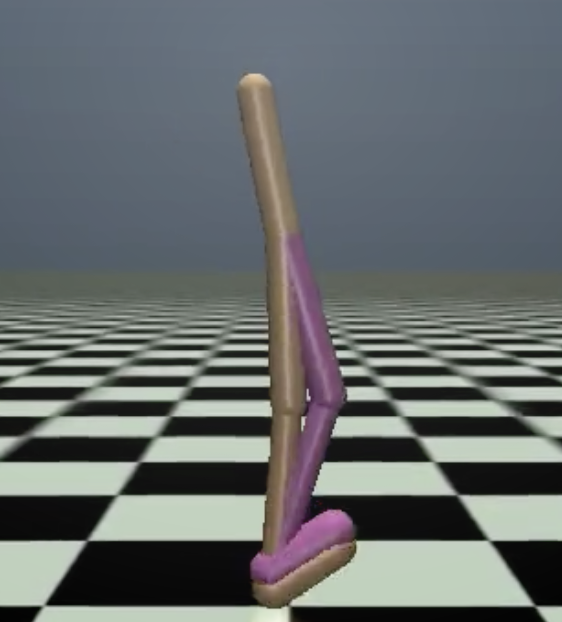}
    \caption{Mujoco environments used for our experiments. Respectively HalfCheetah-v2, Humanoid-v2, Ant-v2, and Walker2d-v2. Articulations are controlled to make them walk. Rewards depends on the covered distance.}
    \label{fig:mujoco}
\end{figure}

To further validate the GRI method we conducted experiments on selected Mujoco \cite{mujoco} environments, shown in \cref{fig:mujoco}. Expert data were generated using chainerrl \cite{chainerrl} pretrained RL agent weights and contain 200,000 samples. For each environment, the value of $r_{demo}$ was chosen as the highest value chainerrl expert agent reached during the generation of the dataset. As we did not find real expert data on Mujoco environments, expert data is not always significantly better than our trained vanilla RL network. Hence, this study assesses the efficiency of GRI even with suboptimal expert data.

\begin{figure*}[h!]
    \centering
    \includegraphics[width=0.43\linewidth]{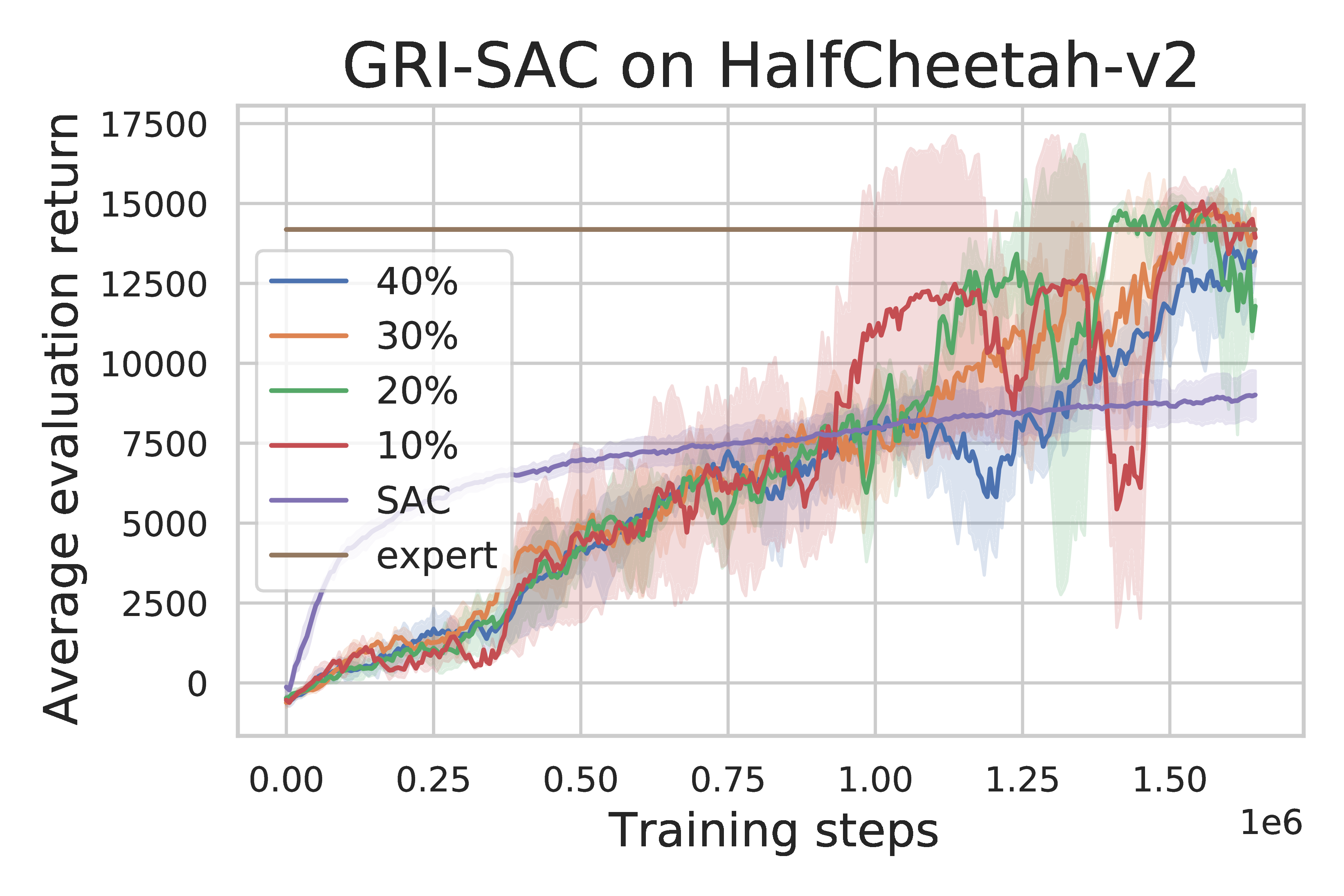}
    \includegraphics[width=0.43\linewidth]{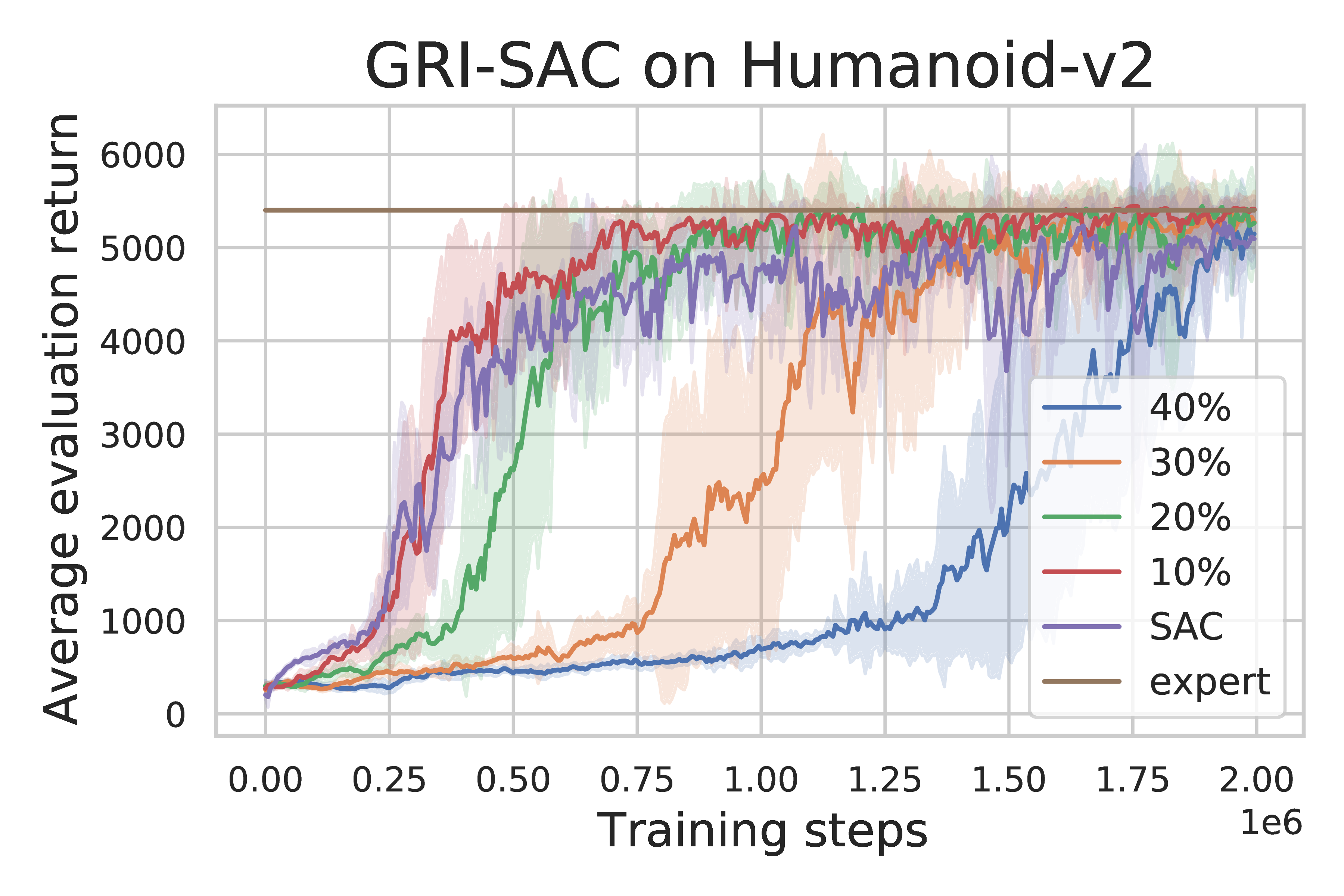}
    \includegraphics[width=0.43\linewidth]{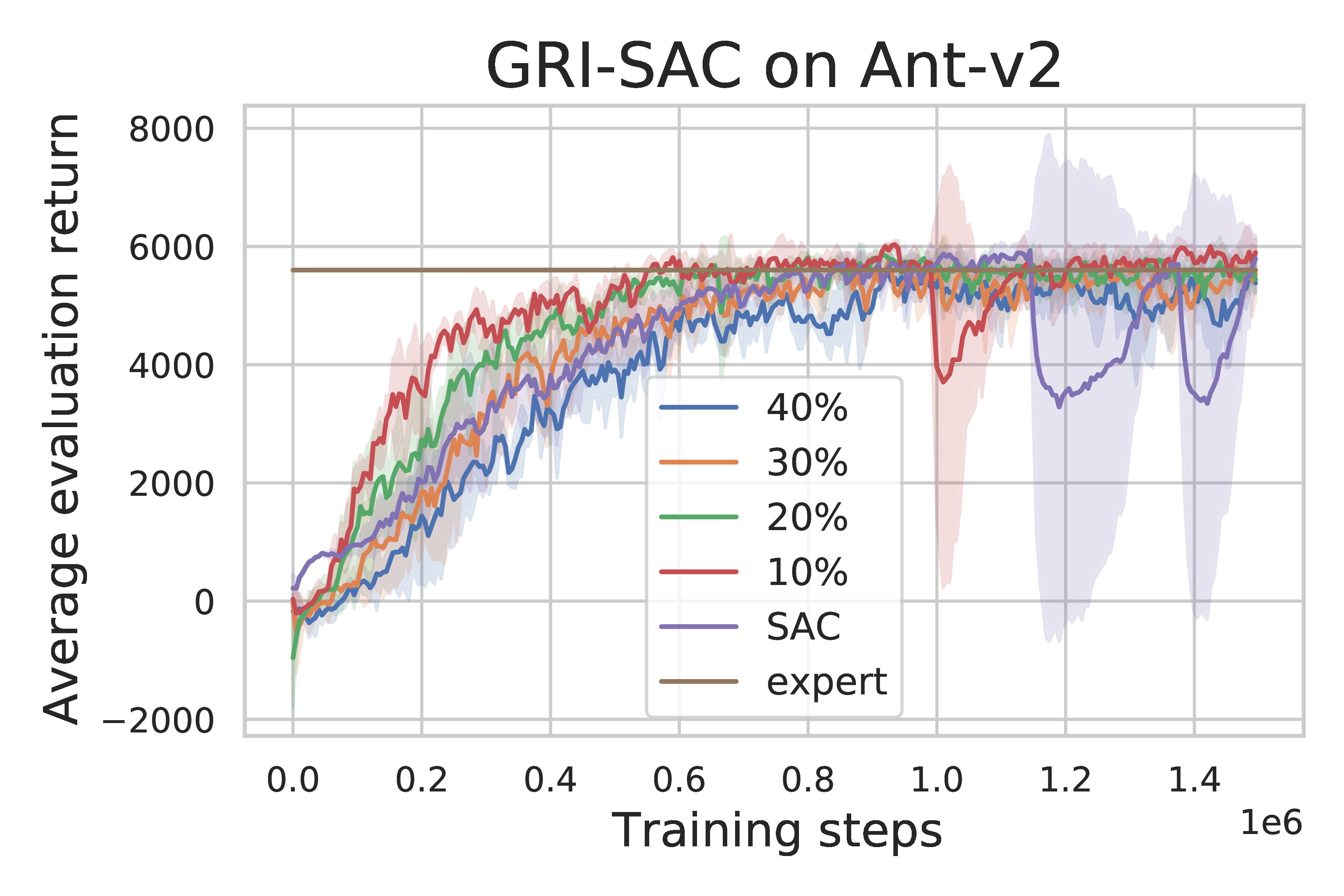}
    \includegraphics[width=0.43\linewidth]{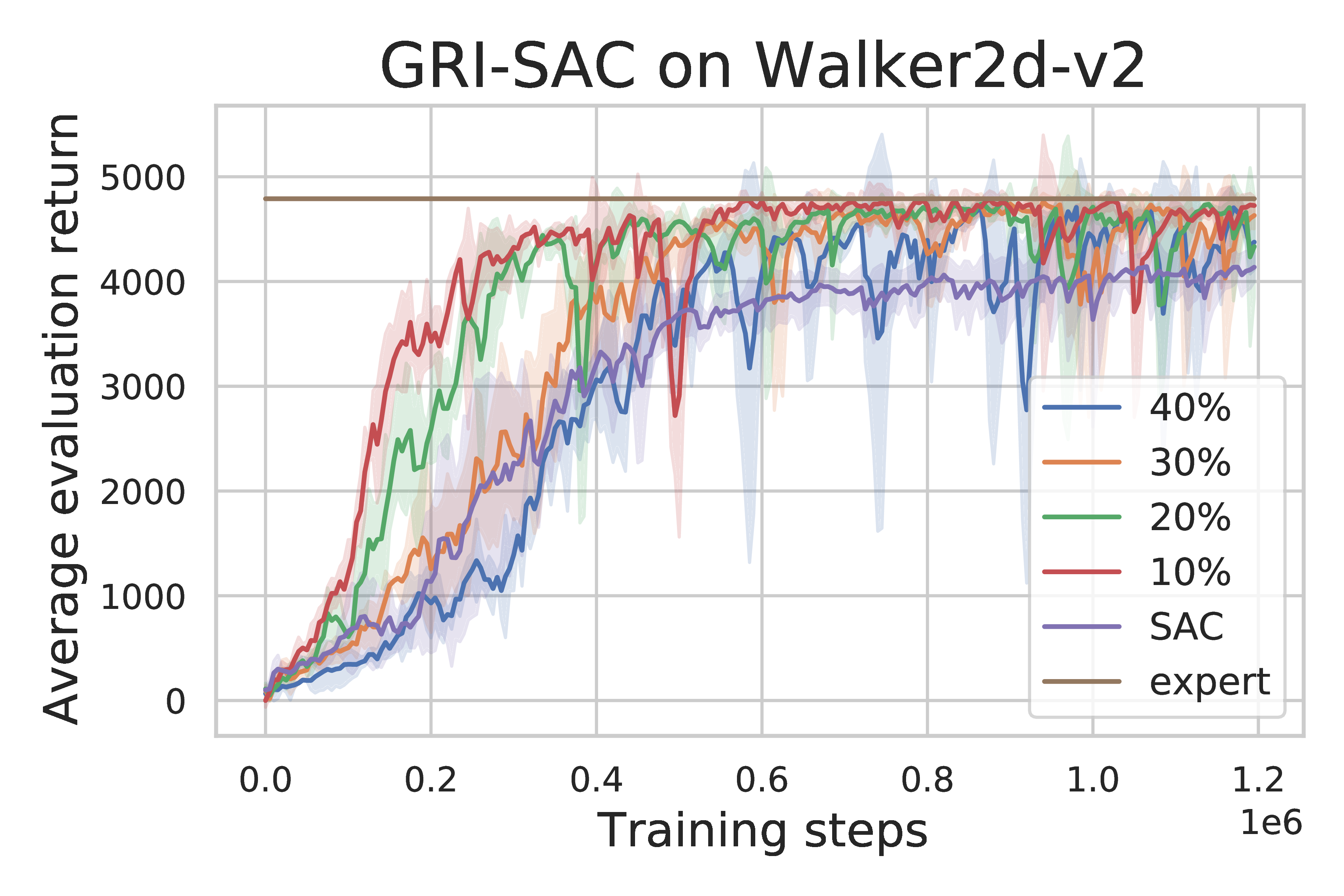}
    \caption{Evolution of the evaluation reward on Mujoco environments with different proportions of demonstration agents with GRI-SAC. GRI-SAC with 0\% demonstration agent is vanilla SAC. We observe that GRI-SAC always reaches the level of the expert even when the expert is significantly better than the trained vanilla SAC. The proportion of demonstration agent have a significant impact on the dynamic of the convergence.}
    \label{fig:num_fake}
\end{figure*}

\textbf{Study on the proportion of demonstration agents.} Experiments being faster on Mujoco environments than on CARLA we were able to investigate the impact of the proportion of demonstration agents.
For these experiments we used a GRI-SAC i.e. a GRI algorithm using SAC \cite{sac} as DRL backbone, and we vary the proportion of demonstration agents between 0\% and 40\%. Each experiment has been repeated three times, with different seeds. \cref{fig:num_fake} presents the results with the variances and the evaluation reward of the expert. Experiments were conducted with public code from GitHub\footnote{Original code from \url{https://github.com/dongminlee94/deep_rl}} which has been adapted with GRI.

We observe three different dynamics. 
\begin{itemize}
    \itemsep0em 
    \item For HalfCheetah-v2, a difficult task on which the expert is significantly stronger than the trained SAC, we observe that the beginning of the training is slower using GRI-SAC; we call this a \textit{warm up phase} which we will explain further in \cref{sec:limitation}. However the rewards turns out to become significantly higher after some time. On this game, GRI-SAC is better than SAC with every proportion of demonstration agents. Best scores were reached with 10\% and 20\% of demonstration agents.
    \item For Humanoid-v2, a difficult task on which the expert is just a little stronger than the trained SAC, we observe that the higher the number of demonstration agents is, the longer the \textit{warm up phase} is. Nonetheless, GRI-SAC models end up having higher rewards after their warm up phase. Best scores are reached with 10\% and 20\% of demonstration agents.
    \item Ant-v2 and Walker2d-v2 are the easiest tasks of the four evaluated. On Ant-v2 the SAC agent reaches the expert level, converging similarly as GRI-SAC regardless of the number of demonstration agents used. Nevertheless, GRI-SAC converges faster with 10\% and 20\% demonstration agents. On Walker2d-v2 the final reward of GRI-SAC is significantly higher and reaches the expert level, while SAC remains below.
\end{itemize}

More experiments were conducted with the proportion of demonstration agents varying between 50\% and 90\%. Results were significantly worse than using 20\% demonstration agents. We therefore conclude that the proportion of demonstration agent should not exceed 50\%. We discuss some qualitative insights in the limitation sections.

These experiments reveals, at least on the evaluated Mujoco environments, that 20\% demonstration agents seems to be the best choice for GRI-SAC to reach the expert level. 

\subsubsection{GRI with DDPG as DRL backbone}
\begin{figure*}[t!]
    \centering
    \includegraphics[width=0.43\linewidth]{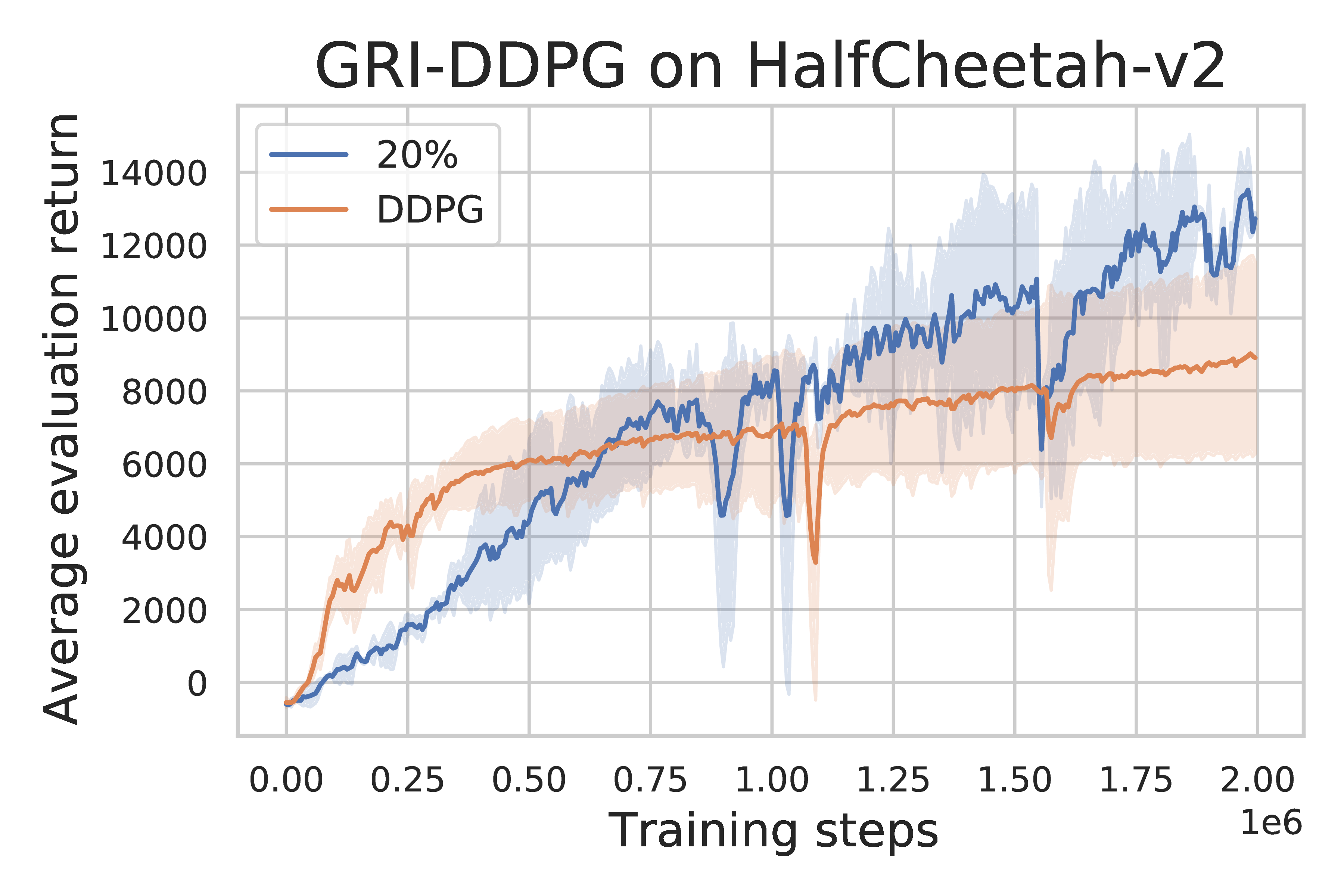}
    \includegraphics[width=0.43\linewidth]{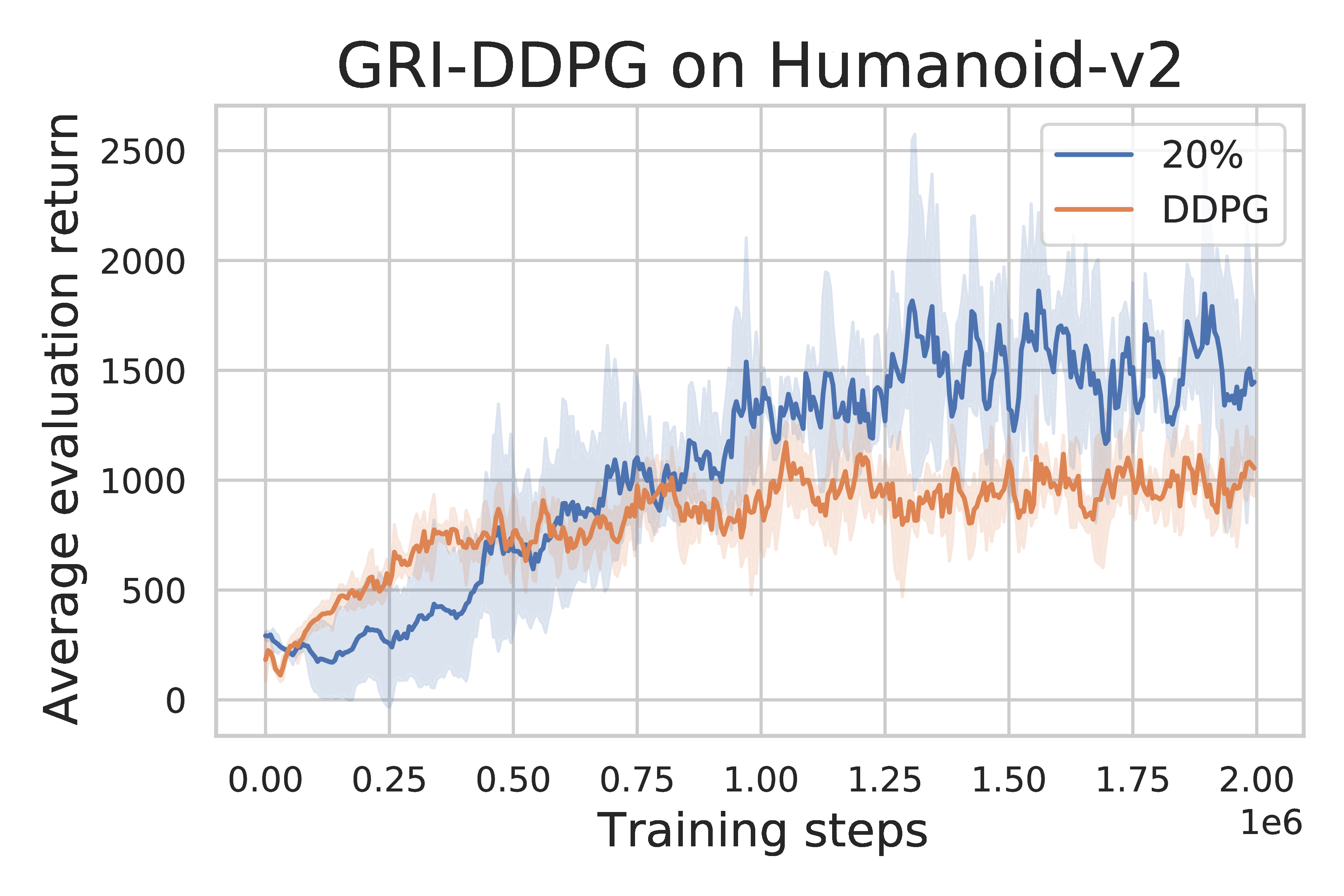}
    \includegraphics[width=0.43\linewidth]{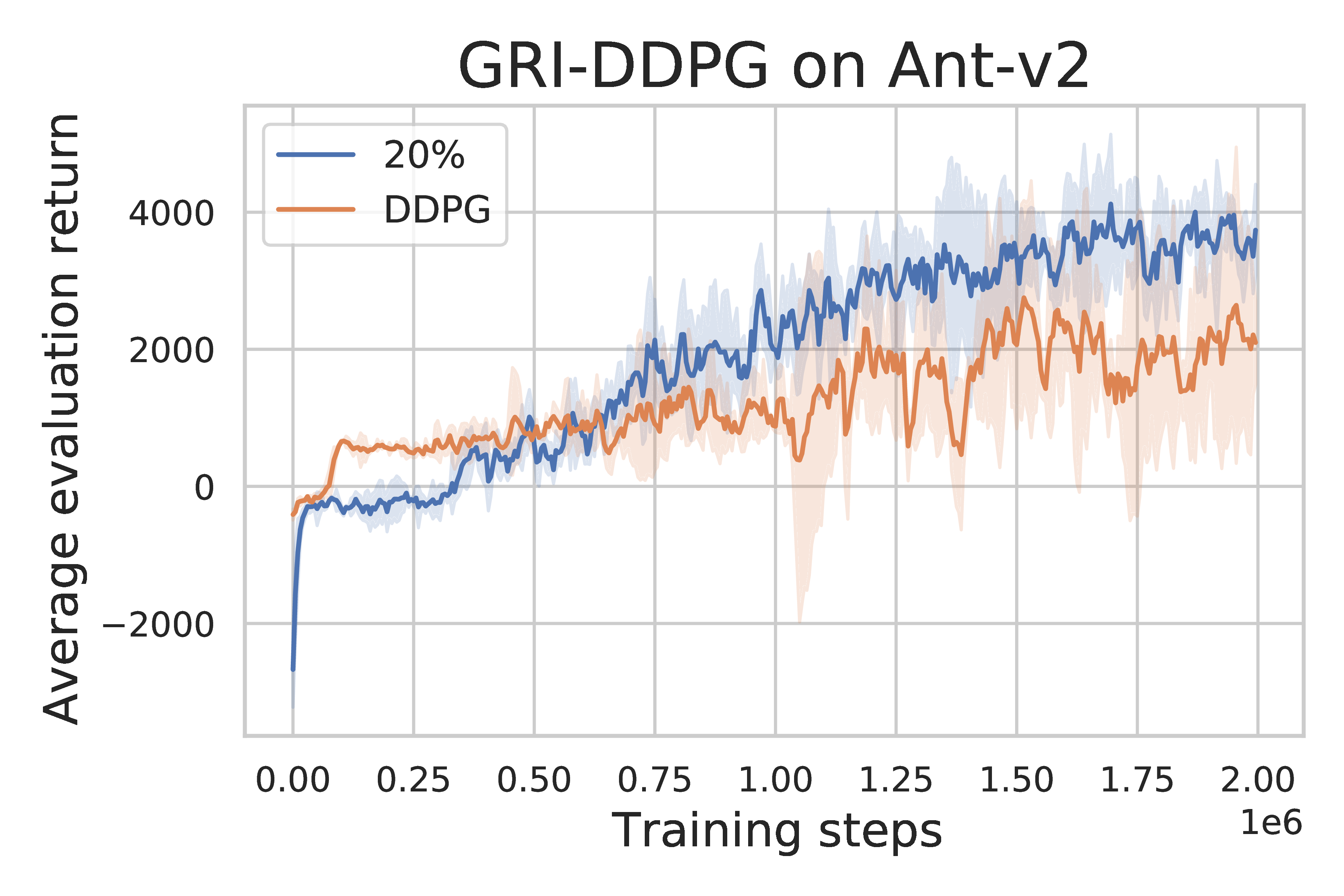}
    \includegraphics[width=0.43\linewidth]{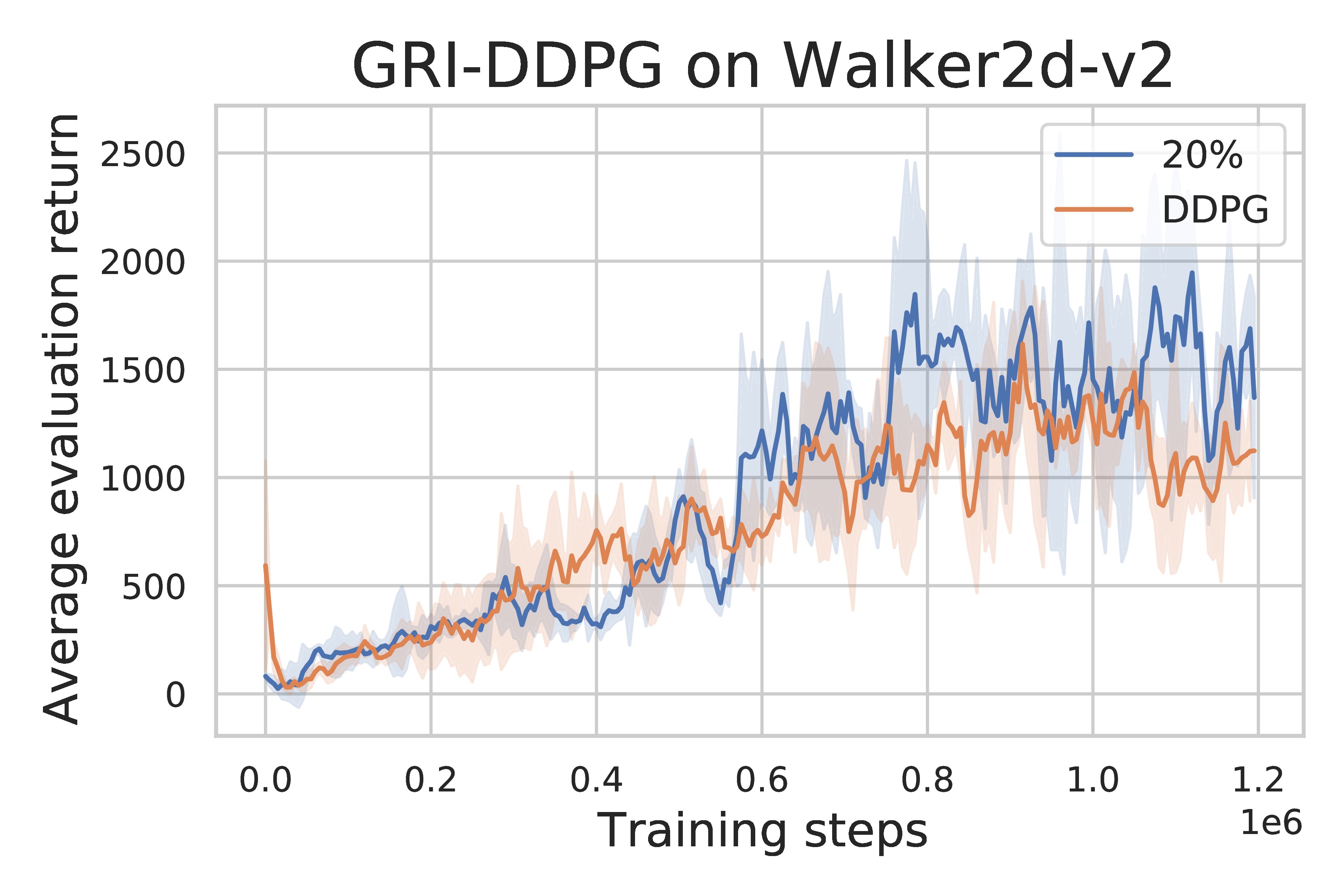}
    \caption{Comparaison of the evaluation reward  evolution on Mujoco environments between  GRI-DDPG with 20\% of demonstration agents and vanilla DDPG. GRI-DDPG systematically leads to a better reward than vanilla DDPG. However, contrarily to GRI-SAC, GRI-DDPG with 20\% demonstration agents does not systematically reach the expert level. }
    \label{fig:xp_ddpgri}
\end{figure*}

We also investigated the contribution of the DRL backbone to assess the generalizability of the GRI method. To do so, we evaluated the same tasks with the Deep Deterministic Policy Gradient (DDPG) algorithm \cite{ddpg} instead of SAC. For these experiments, we fixed the proportion of demonstration agents to 20\%. Results are shown in \cref{fig:xp_ddpgri}.

\addtolength{\textheight}{+0.062cm}   

We observe that, alike to GRI-SAC with a proportion of 20\% demonstration agents, GRI-DDPG reaches better results than DDPG on all the tested environments. However, GRI-DDPG does not systematically reach the level of the expert. While final rewards are better with SAC and GRI-SAC, the dynamics of the rewards evolution is about the same with both backbones, cf. \cref{fig:num_fake}. We can conclude that GRI is easily adaptable and generalizes to locomotion tasks where it robustly outperforms two alternative methods.

\subsection{Limitations}
\label{sec:limitation}
The main limitations of this method are consequences of our initial hypothesis that demonstration data can always be associated with a constant maximal reward $r_{demo}$. 

A first limitation occurs if the demonstration data is not constantly optimal, e.g. due to low expert performance on some aspect of a given task, as this introduces noise in the reward function. This is the case in our demonstration dataset on the CARLA simulator, as expert data have been generated using an imperfect autopilot containing $\sim 10\%$ noisy demonstrations. Still, GRIAD showed to improve our model by a significant margin over vanilla RL. Therefore, we can consider the GRI setup to present some robustness to noisy demonstrations.

A second limitation of our approach is the warm-up phase on some difficult environments, as observed in \cref{fig:num_fake} on HalfCheetah-v2 and Humanoid-v2. This warm up phase can be seen as the consequence of a distribution shift. Indeed, GRI suffers from a sort of distribution shift when the training expert data mostly represent actions made in states not reached yet by the exploration agents. In particular, we observed this effect on HalfCheetah-v2: the expert agent does not walk but jumps as soon as it touches the ground, which is a complex yet highly efficient strategy. But to reach a state where it can successfully jump, it needs some warm up to gain the required speed and momentum by doing some low reward actions. Hence, our GRI-SAC agents learns to jump before it is able to walk, making it fall. Once the agent learned how to reach the jumping state, reward steadily increase until convergence. However, we observe that the lower the proportion of demonstration agents is, the faster the model is able to recover from this distribution shift. Indeed, collecting more exploration data following the current agent policy compensates the distribution shift between demonstration and exploration data. 

Finally, a third limitation of our approach is the inconsistency of the rewards associated to some common actions collected by both the demonstration and exploration agents. Still on the HalfCheetah-v2 example, the demonstration agent will reward expert actions at the beginning of the agent run with the high demonstration reward, while the  exploration agent will receive poor reward for the same exact actions. This induces a sort of discrepancy between data coming from the offline demonstration agent and experiences coming from the online RL exploration agent. It also implies an overestimation of demonstration actions. However, allocating high reward to demonstration data which are not correlated with the actual reward of the environment might encourage the agent to get to states closer to the expert's ones. Nonetheless, it is difficult to assess the impact on the training in practice.

\section{CONCLUSION}
We present GRI, a method that efficiently leverages both expert demonstrations and environment exploration. GRI is straightforward to implement over any off-policy deep reinforcement learning algorithm. GRI-based algorithms improve data efficiency compared to vanilla reinforcement algorithms and do not suffer from distribution shift as much as imitation learning methods. This method also proved to be robust to noisy demonstrations in the expert dataset. We applied GRI to autonomous driving with the distributed GRIAD algorithm and outperformed the previous camera-based state-of-the-art on the CARLA Leaderboard. Finally, we showed its generalizability using different DRL backbones on several Mujoco continuous control environments and highlighted its robustness. In future work, we plan on focusing on LiDAR and camera fusion for GRIAD, as it recently showed to significantly improve the driving quality on CARLA.









\bibliography{references}

\begin{thebibliography}{10}
\providecommand{\url}[1]{#1}
\csname url@samestyle\endcsname
\providecommand{\newblock}{\relax}
\providecommand{\bibinfo}[2]{#2}
\providecommand{\BIBentrySTDinterwordspacing}{\spaceskip=0pt\relax}
\providecommand{\BIBentryALTinterwordstretchfactor}{4}
\providecommand{\BIBentryALTinterwordspacing}{\spaceskip=\fontdimen2\font plus
\BIBentryALTinterwordstretchfactor\fontdimen3\font minus
  \fontdimen4\font\relax}
\providecommand{\BIBforeignlanguage}[2]{{%
\expandafter\ifx\csname l@#1\endcsname\relax
\typeout{** WARNING: IEEEtran.bst: No hyphenation pattern has been}%
\typeout{** loaded for the language `#1'. Using the pattern for}%
\typeout{** the default language instead.}%
\else
\language=\csname l@#1\endcsname
\fi
#2}}
\providecommand{\BIBdecl}{\relax}
\BIBdecl

\bibitem{bojarski}
\BIBentryALTinterwordspacing
M.~Bojarski, D.~D. Testa, D.~Dworakowski, B.~Firner, B.~Flepp, P.~Goyal, L.~D.
  Jackel, M.~Monfort, U.~Muller, J.~Zhang, X.~Zhang, J.~Zhao, and K.~Zieba,
  ``End to end learning for self-driving cars,'' \emph{CoRR}, vol.
  abs/1604.07316, 2016. [Online]. Available:
  \url{http://arxiv.org/abs/1604.07316}
\BIBentrySTDinterwordspacing

\bibitem{il-bullshit}
T.~Osa, J.~Pajarinen, G.~Neumann, J.~A. Bagnell, P.~Abbeel, and J.~Peters,
  2018.

\bibitem{transfuser}
A.~Prakash, K.~Chitta, and A.~Geiger, ``Multi-modal fusion transformer for
  end-to-end autonomous driving,'' in \emph{Proceedings IEEE Conf. on Computer
  Vision and Pattern Recognition (CVPR)}, 2021.

\bibitem{marin-fisheye}
M.~Toromanoff, E.~Wirbel, F.~Wilhelm, C.~Vejarano, X.~Perrotton, and
  F.~Moutarde, ``End to end vehicle lateral control using a single fisheye
  camera,'' in \emph{2018 IEEE/RSJ International Conference on Intelligent
  Robots and Systems (IROS)}, 2018, pp. 3613--3619.

\bibitem{dqn}
\BIBentryALTinterwordspacing
V.~Mnih, K.~Kavukcuoglu, D.~Silver, A.~A. Rusu, J.~Veness, M.~G. Bellemare,
  A.~Graves, M.~Riedmiller, A.~K. Fidjeland, G.~Ostrovski, S.~Petersen,
  C.~Beattie, A.~Sadik, I.~Antonoglou, H.~King, D.~Kumaran, D.~Wierstra,
  S.~Legg, and D.~Hassabis, ``\BIBforeignlanguage{en}{Human-level control
  through deep reinforcement learning},''
  \emph{\BIBforeignlanguage{en}{Nature}}, vol. 518, no. 7540, pp. 529--533,
  Feb. 2015. [Online]. Available:
  \url{http://www.nature.com/articles/nature14236}
\BIBentrySTDinterwordspacing

\bibitem{ppo}
\BIBentryALTinterwordspacing
J.~Schulman, F.~Wolski, P.~Dhariwal, A.~Radford, and O.~Klimov, ``Proximal
  policy optimization algorithms.'' \emph{CoRR}, vol. abs/1707.06347, 2017.
  [Online]. Available:
  \url{http://dblp.uni-trier.de/db/journals/corr/corr1707.html#SchulmanWDRK17}
\BIBentrySTDinterwordspacing

\bibitem{td3}
\BIBentryALTinterwordspacing
S.~Fujimoto, H.~van Hoof, and D.~Meger, ``Addressing function approximation
  error in actor-critic methods,'' in \emph{Proceedings of the 35th
  International Conference on Machine Learning}, ser. Proceedings of Machine
  Learning Research, J.~Dy and A.~Krause, Eds., vol.~80.\hskip 1em plus 0.5em
  minus 0.4em\relax PMLR, 10--15 Jul 2018, pp. 1587--1596. [Online]. Available:
  \url{https://proceedings.mlr.press/v80/fujimoto18a.html}
\BIBentrySTDinterwordspacing

\bibitem{a3c}
\BIBentryALTinterwordspacing
V.~Mnih, A.~P. Badia, M.~Mirza, A.~Graves, T.~Lillicrap, T.~Harley, D.~Silver,
  and K.~Kavukcuoglu, ``Asynchronous methods for deep reinforcement learning,''
  in \emph{Proceedings of The 33rd International Conference on Machine
  Learning}, ser. Proceedings of Machine Learning Research, M.~F. Balcan and
  K.~Q. Weinberger, Eds., vol.~48.\hskip 1em plus 0.5em minus 0.4em\relax New
  York, New York, USA: PMLR, 20--22 Jun 2016, pp. 1928--1937. [Online].
  Available: \url{https://proceedings.mlr.press/v48/mniha16.html}
\BIBentrySTDinterwordspacing

\bibitem{carla}
A.~Dosovitskiy, G.~Ros, F.~Codevilla, A.~Lopez, and V.~Koltun, ``{CARLA}: {An}
  open urban driving simulator,'' in \emph{Proceedings of the 1st Annual
  Conference on Robot Learning}, 2017, pp. 1--16, license: CC-BY.

\bibitem{wor}
D.~Chen, V.~Koltun, and P.~Kr{\"a}henb{\"u}hl, ``Learning to drive from a world
  on rails,'' in \emph{ICCV}, 2021.

\bibitem{nocrash}
\BIBentryALTinterwordspacing
F.~Codevilla, E.~Santana, A.~Lopez, and A.~Gaidon, ``Exploring the
  {Limitations} of {Behavior} {Cloning} for {Autonomous} {Driving},'' in
  \emph{2019 {IEEE}/{CVF} {International} {Conference} on {Computer} {Vision}
  ({ICCV})}.\hskip 1em plus 0.5em minus 0.4em\relax Seoul, Korea (South): IEEE,
  Oct. 2019, pp. 9328--9337. [Online]. Available:
  \url{https://ieeexplore.ieee.org/document/9009463/}
\BIBentrySTDinterwordspacing

\bibitem{mujoco}
E.~Todorov, T.~Erez, and Y.~Tassa, ``Mujoco: A physics engine for model-based
  control,'' in \emph{2012 IEEE/RSJ International Conference on Intelligent
  Robots and Systems}, 2012, pp. 5026--5033.

\bibitem{lbc}
D.~Chen, B.~Zhou, V.~Koltun, and P.~Kr\"ahenb\"uhl, ``Learning by cheating,''
  in \emph{Conference on Robot Learning (CoRL)}, 2019.

\bibitem{splitnet}
\BIBentryALTinterwordspacing
D.~Gordon, A.~Kadian, D.~Parikh, J.~Hoffman, and D.~Batra, ``{SplitNet}:
  {Sim2Sim} and {Task2Task} {Transfer} for {Embodied} {Visual} {Navigation},''
  in \emph{2019 {IEEE}/{CVF} {International} {Conference} on {Computer}
  {Vision} ({ICCV})}.\hskip 1em plus 0.5em minus 0.4em\relax Seoul, Korea
  (South): IEEE, Oct. 2019, pp. 1022--1031. [Online]. Available:
  \url{https://ieeexplore.ieee.org/document/9009082/}
\BIBentrySTDinterwordspacing

\bibitem{architecture-marin}
M.~Toromanoff, E.~Wirbel, and F.~Moutarde, ``End-to-end model-free
  reinforcement learning for urban driving using implicit affordances,'' in
  \emph{Proceedings of the IEEE/CVF Conference on Computer Vision and Pattern
  Recognition (CVPR)}, June 2020.

\bibitem{roach}
Z.~Zhang, A.~Liniger, D.~Dai, F.~Yu, and L.~Van~Gool, ``End-to-end urban
  driving by imitating a reinforcement learning coach,'' in \emph{Proceedings
  of the IEEE/CVF International Conference on Computer Vision (ICCV)}, 2021.

\bibitem{dqfd}
\BIBentryALTinterwordspacing
T.~Hester, M.~Vecer{\'{\i}}k, O.~Pietquin, M.~Lanctot, T.~Schaul, B.~Piot,
  A.~Sendonaris, G.~Dulac{-}Arnold, I.~Osband, J.~P. Agapiou, J.~Z. Leibo, and
  A.~Gruslys, ``Learning from demonstrations for real world reinforcement
  learning,'' \emph{CoRR}, vol. abs/1704.03732, 2017. [Online]. Available:
  \url{http://arxiv.org/abs/1704.03732}
\BIBentrySTDinterwordspacing

\bibitem{sqil}
\BIBentryALTinterwordspacing
S.~Reddy, A.~D. Dragan, and S.~Levine, ``{SQIL:} imitation learning via
  regularized behavioral cloning,'' \emph{CoRR}, vol. abs/1905.11108, 2019.
  [Online]. Available: \url{http://arxiv.org/abs/1905.11108}
\BIBentrySTDinterwordspacing

\bibitem{dapg}
\BIBentryALTinterwordspacing
A.~Rajeswaran, V.~Kumar, A.~Gupta, J.~Schulman, E.~Todorov, and S.~Levine,
  ``Learning complex dexterous manipulation with deep reinforcement learning
  and demonstrations,'' \emph{CoRR}, vol. abs/1709.10087, 2017. [Online].
  Available: \url{http://arxiv.org/abs/1709.10087}
\BIBentrySTDinterwordspacing

\bibitem{sacr2}
J.~B. Martin, R.~Chekroun, and F.~Moutarde, ``Learning from demonstrations with
  sacr2: Soft actor-critic with reward relabeling,'' \emph{arXiv preprint
  arXiv:2110.14464}, 2021.

\bibitem{nac}
\BIBentryALTinterwordspacing
D.~Xu, S.~Nair, Y.~Zhu, J.~Gao, A.~Garg, L.~Fei-Fei, and S.~Savarese, ``Neural
  {Task} {Programming}: {Learning} to {Generalize} {Across} {Hierarchical}
  {Tasks},'' in \emph{2018 {IEEE} {International} {Conference} on {Robotics}
  and {Automation} ({ICRA})}.\hskip 1em plus 0.5em minus 0.4em\relax Brisbane,
  QLD: IEEE, May 2018, pp. 3795--3802. [Online]. Available:
  \url{https://ieeexplore.ieee.org/document/8460689/}
\BIBentrySTDinterwordspacing

\bibitem{rl-id}
\BIBentryALTinterwordspacing
Y.~Gao, H.~Xu, J.~Lin, F.~Yu, S.~Levine, and T.~Darrell, ``Reinforcement
  learning from imperfect demonstrations,'' \emph{CoRR}, vol. abs/1802.05313,
  2018. [Online]. Available: \url{http://arxiv.org/abs/1802.05313}
\BIBentrySTDinterwordspacing

\bibitem{sac}
\BIBentryALTinterwordspacing
T.~Haarnoja, A.~Zhou, P.~Abbeel, and S.~Levine, ``Soft actor-critic: Off-policy
  maximum entropy deep reinforcement learning with a stochastic actor,'' 2018.
  [Online]. Available: \url{https://openreview.net/forum?id=HJjvxl-Cb}
\BIBentrySTDinterwordspacing

\bibitem{ddpg}
\BIBentryALTinterwordspacing
T.~P. Lillicrap, J.~J. Hunt, A.~Pritzel, N.~Heess, T.~Erez, Y.~Tassa,
  D.~Silver, and D.~Wierstra, ``Continuous control with deep reinforcement
  learning,'' in \emph{4th International Conference on Learning
  Representations, {ICLR} 2016, San Juan, Puerto Rico, May 2-4, 2016,
  Conference Track Proceedings}, Y.~Bengio and Y.~LeCun, Eds., 2016. [Online].
  Available: \url{http://arxiv.org/abs/1509.02971}
\BIBentrySTDinterwordspacing

\bibitem{rainbow}
\BIBentryALTinterwordspacing
M.~Hessel, J.~Modayil, H.~van Hasselt, T.~Schaul, G.~Ostrovski, W.~Dabney,
  D.~Horgan, B.~Piot, M.~G. Azar, and D.~Silver, ``Rainbow: Combining
  improvements in deep reinforcement learning,'' \emph{CoRR}, vol.
  abs/1710.02298, 2017. [Online]. Available:
  \url{http://arxiv.org/abs/1710.02298}
\BIBentrySTDinterwordspacing

\bibitem{iqn}
\BIBentryALTinterwordspacing
W.~Dabney, G.~Ostrovski, D.~Silver, and R.~Munos, ``Implicit quantile networks
  for distributional reinforcement learning,'' in \emph{Proceedings of the 35th
  International Conference on Machine Learning}, ser. Proceedings of Machine
  Learning Research, J.~Dy and A.~Krause, Eds., vol.~80.\hskip 1em plus 0.5em
  minus 0.4em\relax PMLR, 10--15 Jul 2018, pp. 1096--1105. [Online]. Available:
  \url{https://proceedings.mlr.press/v80/dabney18a.html}
\BIBentrySTDinterwordspacing

\bibitem{efficientnet}
\BIBentryALTinterwordspacing
M.~Tan and Q.~Le, ``{E}fficient{N}et: Rethinking model scaling for
  convolutional neural networks,'' in \emph{Proceedings of the 36th
  International Conference on Machine Learning}, ser. Proceedings of Machine
  Learning Research, K.~Chaudhuri and R.~Salakhutdinov, Eds., vol.~97.\hskip
  1em plus 0.5em minus 0.4em\relax PMLR, 09--15 Jun 2019, pp. 6105--6114.
  [Online]. Available: \url{https://proceedings.mlr.press/v97/tan19a.html}
\BIBentrySTDinterwordspacing

\bibitem{drl-marin}
\BIBentryALTinterwordspacing
M.~Toromanoff, E.~Wirbel, and F.~Moutarde, ``{Is Deep Reinforcement Learning
  Really Superhuman on Atari?}'' in \emph{{Deep Reinforcement Learning Workshop
  of 39th Conference on Neural Information Processing Systems (Neurips'2019)}},
  Vancouver, Canada, Dec. 2019. [Online]. Available:
  \url{https://hal-mines-paristech.archives-ouvertes.fr/hal-02368263}
\BIBentrySTDinterwordspacing

\bibitem{chainerrl}
\BIBentryALTinterwordspacing
Y.~Fujita, P.~Nagarajan, T.~Kataoka, and T.~Ishikawa, ``Chainerrl: A deep
  reinforcement learning library,'' \emph{Journal of Machine Learning
  Research}, vol.~22, no.~77, pp. 1--14, 2021. [Online]. Available:
  \url{http://jmlr.org/papers/v22/20-376.html}
\BIBentrySTDinterwordspacing

\end{thebibliography}
\bibliographystyle{IEEEtran}
\end{document}